\pdfoutput=1

\documentclass[11pt]{article}

\usepackage[preprint]{acl}

\usepackage{times}
\usepackage{latexsym}
\usepackage{booktabs}

\usepackage[T1]{fontenc}

\usepackage[utf8]{inputenc}

\usepackage{microtype}

\usepackage{inconsolata}

\usepackage{graphicx}

\usepackage{amsmath}
\usepackage{amssymb}
\usepackage{mathtools}
\usepackage{amsthm}
\usepackage[many]{tcolorbox}    	

\usepackage{algorithm}
\usepackage{algpseudocode}

\usepackage{mdframed}

\title{Language-Informed Synthesis of Rational Agent Models \\ for Grounded Theory-of-Mind Reasoning On-The-Fly}

\author{
  Lance Ying$^{1,2}$, \quad
  Ryan Truong$^2$,\quad
  Katherine M. Collins$^3$,\quad
  Cedegao E. Zhang$^1$,\\
  {\bf
  Megan Wei$^4$, \quad
  Tyler Brooke-Wilson$^{5}$, \quad
  Tan Zhi-Xuan$^{1\ddagger}$, \quad
  Lionel Wong$^{6\ddagger}$,
  }\\
    {\bf
  Joshua B. Tenenbaum$^{1\ddagger}$
  }
  \vspace{0.2cm}
   \\ 
  $^1$MIT  \quad$^2$Harvard University  \quad$^3$University of Cambridge \quad \\ $^4$Brown University  \quad $^5$Yale University  \quad $^6$Stanford University \\ 
 $^\ddagger$ co-senior authors
}

\begin{document}
\maketitle

\begin{abstract}

Drawing real world social inferences usually requires taking into account information from multiple modalities. Language is a particularly powerful source of information in social settings, especially in novel situations where language can provide both abstract information about the environment dynamics and concrete specifics about an agent that cannot be easily visually observed. In this paper, we propose Language-Informed Rational Agent Synthesis (LIRAS), a framework for drawing context-specific social inferences that integrate linguistic and visual inputs. LIRAS frames multimodal social reasoning as a process of constructing structured but situation-specific agent and environment representations – leveraging multimodal language models to parse language and visual inputs into unified symbolic representations, over which a Bayesian inverse planning engine can be run to produce granular probabilistic judgments. On a range of existing and new social reasoning tasks derived from cognitive science experiments, we find that our model (instantiated with a comparatively lightweight VLM) outperforms ablations and state-of-the-art models in capturing human judgments across all domains.

\end{abstract}

\section{Introduction}
Making sense of any real social situation requires integrating many different sources of information. \textit{Language}, in particular, can fundamentally recast our understanding of the social environment around us. Being told the abstract dynamics underlying a social institution, from the rules of American football to the norms of drive-through restaurants, gives us a useful overarching picture of people’s goals and intentions in unfamiliar settings. Other times, language can provide specifics about particular people and environments. Hearing that a friend tends to get hungry around midnight and keeps a spare stash of chocolate in the highest pantry shelf, for instance, gives new meaning to a few glimpses of someone bumbling around the kitchen in the dark. These tidbits of socially-relevant information from language allow us to draw much richer, more flexible, and often quite situation-dependent conclusions about the behavior we see. 

How do we integrate language with perceptual information to support this kind of \textit{grounded} but often\textit{ highly ad-hoc} social reasoning, in which language can flexibly restructure our inferences about the agents we observe? This setting poses particularly acute challenges for two dominant flavors of computational work in social reasoning. For approaches that cast social reasoning as principled inferences over structured models of agents and environments, as in many symbolic AI and cognitive science frameworks (e.g., \citealt{baker2011bayesian,jara2016naive}), this setting tests the scalability and breadth that can be achieved using (usually hand-engineered) symbolic models of particular environments and domains. For approaches that use large-scale neural models trained on language and visual inputs, like recent LLM and VLM-based systems (e.g., \citealt{kosinski2024evaluating,jin2024mmtom}), this setting tests the generalizability of complex decision-making and latent inference over particularly novel inputs. Ongoing evaluations suggest that \textit{each} component of grounded, ad-hoc social reasoning poses challenges for both dominant approaches \cite{hu2025re,schulze2025visual,jin2024mmtom} – challenges that only compound in the harder multi-modal setting. 

\begin{figure*}[ht!]
    \centering
    \includegraphics[width=1\linewidth]{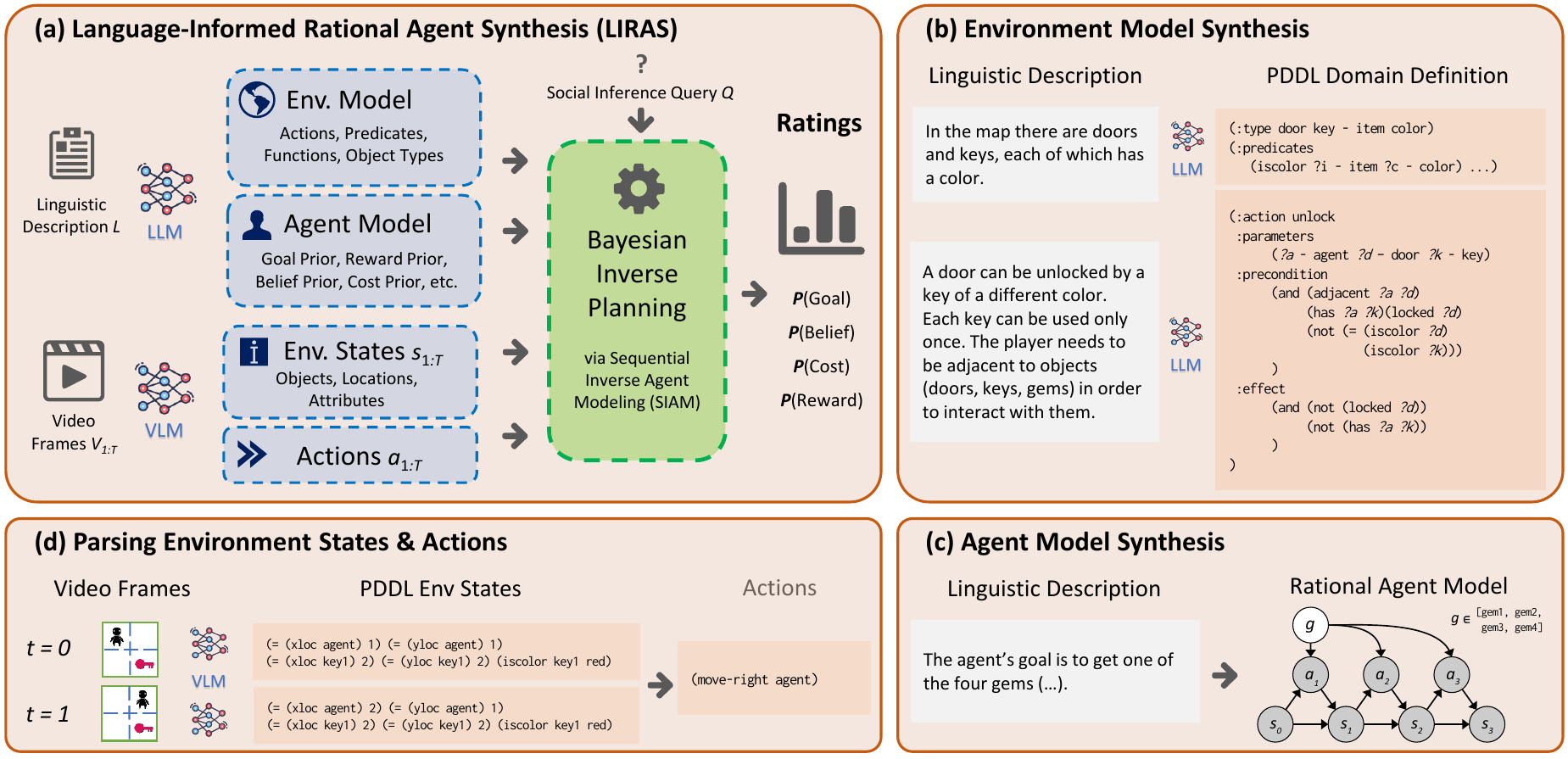}
    \caption{Overview of Language-Informed Rational Agent Synthesis (LIRAS). \textbf{(a)} Overall architecture. \textbf{(b)} LIRAS uses a vision-capable large language model (LLM) to parse linguistic descriptions into a domain-specific environment model expressed in PDDL, which specifies object types, predicates, functions and action definitions. \textbf{(c)} LIRAS constructs a rational agent model based on the linguistic descriptions about the agent, including the space of the agent's goals, beliefs, etc. \textbf{(d)} LIRAS parses visual inputs into symbolic PDDL environment states, and derives agent actions from those states.}
    \label{fig:overview}
\end{figure*}

In this paper, we develop \textbf{Language-Informed Rational Agent Synthesis (LIRAS)}, a framework that can \textbf{integrate linguistic and visual inputs} to draw \textbf{ad-hoc, probabilistic inferences} about agents’ mental states in grounded settings. Our approach achieves this by using neural models to parse language and visual inputs into unified symbolic representations, which support automatic Bayesian inference and inverse planning. Importantly, our approach casts multimodal social reasoning as a process of constructing small but structured environment representations \textit{on-the-fly}, tailored to a particular set of inputs. Rather than hand-construct domain-specific world models, or seek to derive universal features for visual understanding, our approach lets new, relevant details from language change how we parse and reason about any particular visual scene. 

We evaluate our approach on a suite of popular domains used for social cognition experiments, as well as novel variants designed to probe people and models’ generalization capacities under new environment dynamics and abstract rule structures. We demonstrate that our framework, even when built with a lightweight VLM, can capture human-like social reasoning in a context-sensitive way across each of these domains. In contrast, we find that much larger state-of-the-art VLMs, such as OpenAI's o3 model, often fail to reliably integrate linguistic details with visual observations and generally are much more uneven in capturing human judgments across these domains. 

\section{Language-Informed Rational Agent Synthesis (LIRAS)}
\label{sec:liras}

In this paper, we consider how to solve ad-hoc social reasoning tasks that are grounded in visual observations of an agent taking actions over time. Each task $(L, Q, V_{1:T})$ is defined by a linguistic description $L$ of the agent and its environment, a social inference query $Q$ expressed in natural language, and a sequence of $T+1$ video frames $V_{0:T}$ showing how the agent interacts with the environment over time. Given $(L, Q, V_{0:T})$, LIRAS produces $k \geq 1$ graded ratings $R_{1:k} \in \mathbb{R}^k$ about the agent's mental states (e.g. goals, beliefs, etc.) based on the query $Q$. Since probabilistic social reasoning lacks a ``ground-truth'' answer \cite{baker2017rational,ying2025benchmarking}, we assess task performance based on how \emph{human-like} the ratings $R_{1:k}$ are, measuring correlation with a dataset of collected human responses (see Section \ref{sec:experiments}).

To solve these tasks, we propose \emph{Language-Informed Rational Agent Synthesis} (LIRAS, Figure \ref{fig:overview}). LIRAS differs from social reasoning methods which rely on either prompting a single LLM \cite{sap2022neural,moghaddam2023boosting} or scaffolding of LLM calls within a structured probabilistic inference procedure \cite{cross2024hypothetical,kim2025hypothesis,zhang2025autotom}. Instead, given a language description $L$, LIRAS synthesizes a \emph{rational model of an agent and its environment} --- which assigns likelihoods to an agents' actions by solving the (Partially Observable) Markov Decision Process ((PO)MDP) \cite{bellman1958dynamic,kaelbling1998planning} that describes the agent's decision problem --- and draws probabilistic inferences about mental states from the agent's actions via \emph{Bayesian inverse planning} with respect to that agent model \cite{baker2009action,baker2017rational,zhi2020online}. This process is decomposed into four components: (i) \emph{synthesis of an environment model} (Fig. \ref{fig:overview}b); (ii) \emph{synthesis of a rational agent model} (Fig. \ref{fig:overview}c); (iii) \emph{parsing environment states and agent actions} (Fig. \ref{fig:overview}d); (iv) \emph{mental state inference via Sequential Inverse Agent Modeling (SIAM)} (Fig. \ref{fig:overview}a, right), a flexible engine for Bayesian inverse planning that extends Sequential Inverse Plan Search \cite{zhi2020online}. We describe each component below. 



\subsection{Synthesizing Environment Models}
\label{sec:synthesizing-env-models}

In order to synthesize a rational agent model, we first need to synthesize a \emph{environment model} or \emph{world model} that the agent model is situated within. Following work that translates language into symbolic world models represented as programs \cite{wong2023word,tang2024worldcoder,wong2024learning,liu2023llmp,xie2023translating}, LIRAS achieves this by using an LLM to translate the task description $L$ into a \emph{planning domain} $\mathcal{D}$ represented in the Planning Domain Definition Language (PDDL)~\cite{aeronautiques1998pddl,zhi2022pddl}. Given $L$ and an instruction prompt $I_\text{env}$, we rejection sample from the LLM to ensure syntactic and semantic validity of the generated PDDL domain:
{
\setlength{\abovedisplayskip}{4pt}
\setlength{\belowdisplayskip}{4pt}
\begin{equation}
    \mathcal{D} \sim P_\text{LLM}(\mathcal{D}| L, I_\text{env}, \mathcal{D} \text{ is valid})
\end{equation}
}
As shown in Figure \ref{fig:overview}(b), a planning domain $\mathcal{D} = (\mathcal{T}, \mathcal{P}, \mathcal{A})$ consists of a set of object types $\mathcal{T}$, predicates $\mathcal{P}$, and action templates $\mathcal{A}$. Given a specific set of (typed) objects $\mathcal{O}$, a planning domain defines a concrete environment $\mathcal{E} = (\mathsf{S}, \mathsf{A}, P_s)$, where $\mathsf{S}$ is the set of possible environment states formed from predicates defined over objects in $\mathcal{O}$, $\mathsf{A}$ is the set of possible actions derived by filling in action templates with object arguments and $P_s(s_t | s_{t-1}, a_t)$ is an environment transition distribution. This gives us the basic structure on top of which we can define an agent model and perform visual parsing of environment states.

\subsection{Synthesizing Rational Agent Models}
\label{sec:synthesizing-agent-models}

{
When people reason about the mental states of another agent, we form an implicit model of that agent that predicts and explains their actions in light of their goals, beliefs, and other mental states \cite{dennett1981intentional}. Importantly, we assume that the agent is approximately \emph{rational} --- their \emph{beliefs are consistent} with what they observe, and they take \emph{efficient actions} to achieve their goals and satisfy their desires. To formalize this, we follow work in Bayesian theory-of-mind \cite{baker2017rational,jara2016naive,alanqary2021modeling,ying2025understanding}, treating an agent as a generative processes of the following form: 
\setlength{\abovedisplayskip}{4pt}
\setlength{\belowdisplayskip}{4pt}
\begin{alignat}{2}
{\small\textit{Mental Prior:}}& \  m_0 \sim P_{\theta_0}(m_0; s_0) \label{eq:mental-prior} \\
{\small\textit{Mental Update:}}& \  m_t \sim P_{\theta_m}(m_t | s_{t-1}, m_{t-1}) \label{eq:belief-update} \\
{\small\textit{Action Selection:}}& \  a_t \sim P_{a}(a_t | m_t, s_{t-1}) \label{eq:action-selection}
\end{alignat}
Here, $m_t$ represents the mental state of the agent at step $t$, which can include beliefs $b_t$, goals $g_t$, rewards $r_t$ that the agent assigns to achieving a goal, or the perceived costs $c_t$ of certain actions. At each step $t$, the agent may update their beliefs $b_t \in m_t$ based on their observations in $s_{t-1}$ in a way that preserves \emph{consistency}: If some predicate $p$ is observed to hold true in state $s_{t-1}$, then $p$ must also hold true in the updated belief $b_t$. They then take an action $a_t$ to \emph{efficiently} achieve the goals or rewards specified in $m_t$ while minimizing costs. Specifically, we assume that $a_t$ follows a Boltzmann-rational distribution:
\begin{equation}
P_{a}(a_t | m_t, s_{t-1}) \propto \exp\left(\hat Q_{m_t}(s_{t-1}, a_t) \right) 
\end{equation}
where $\hat Q_m(s_{t-1}, a_t)$ is an estimate of the \emph{expected utility} of reaching any goal in mental state $m_t$ by taking action $a_t$ at state $s$. Action $a_t$ then causes a change in the environment $s_t \sim P_s(s_t | s_{t-1}, a_t)$ per the environment model described earlier, and the process repeats.
}

{
Depending on the situation, an observer may not need to model all of the agent's mental states. For instance, if the agent has full observability of the environment $s_t$, there is no need to represent the agents' belief state $b_t$, since $b_t$ will always agree with $s_t$. As such, LIRAS constructs agent models in an \emph{ad-hoc} manner: Given the language description $L$, (and an instruction prompt $I_\text{agent}$), we use an LLM to synthesize the parameters $\Theta = (\theta_0, \theta_m)$ that define the agent model (Figure \ref{fig:overview}(c)):
\setlength{\abovedisplayskip}{4pt}
\setlength{\belowdisplayskip}{4pt}
\begin{equation}
    \Theta \sim P_\text{LLM}(\Theta| L, I_\text{agent})
\end{equation}
$\theta_0$ includes information like the space of possible initial beliefs, goals, goal rewards, or action costs (over which we assume a uniform prior), and $\theta_m$ includes information like the observability of various objects in state $s_t$ (since this affects how the agent updates their beliefs). By flexibly synthesizing different models based on the social situation described in $L$, LIRAS captures human-like flexibility in social reasoning, while preserving the rationality assumptions of belief consistency and action efficiency described above.
}



\subsection{Parsing Environment States and Actions}
\label{sec:parsing-states-and-actions}

To infer an agent's mental states, LIRAS first needs to parse the images $I_{0:T}$ into a sequence of symbolically represented environment states $s_{0:T}$ and actions $a_{1:T}$ (Figure \ref{fig:overview}(d)). We achieve this using a vision-language model (VLM) in a decoding procedure that exploits the grid-based of our environments.\footnote{This is not key to our architecture, but simply what we found was necessary given the current unreliability of VLMs and parsing full 2D visual scenes.} Specifically, for each grid cell in a video frame $V_t$, we use the VLM to detect the object in that cell (if any), and generate corresponding PDDL code for the object's properties. This allows us to parse a state $s_t$ from frame $V_t$:
{
\setlength{\abovedisplayskip}{4pt}
\setlength{\belowdisplayskip}{4pt}
\begin{equation}
    s_t \sim P_\text{VLM}(s_t | V_{t}, I_\text{states})
\end{equation}
}
where $I_\text{states}$ in an instruction prompt. Given the full sequence of environment states $s_{1:t}$, we can then greedily reconstruct each action $a_t$ under the environment model $P_s(s_{t}|s_{t-1},a_t)$:
{
\setlength{\abovedisplayskip}{4pt}
\setlength{\belowdisplayskip}{4pt}
\begin{equation}
    a_t = \operatorname{argmax}_a P_s(s_{t}|s_{t-1}, a)
\end{equation}
}


\subsection{Bayesian Mental State Inference via Sequential Inverse Agent Modeling}
\label{sec:siam}

Having synthesized an environment model, agent model, environment states $s_{1:T}$ and actions $a_{1:T}$, LIRAS answers the social inference queries in $Q$ via \emph{Bayesian inverse planning}. Specifically, we use Sequential Inverse Agent Modeling (SIAM), an extension of the SIPS probabilistic programming architecture for inverse planning \cite{zhi2020online,zhi2024infinite} that supports joint inference over not just goals $g$ (as in \citet{zhi2020online}), but also goal preferences/rewards $r$ (similar to \citet{zhi2022solving}), beliefs (as in belief-space SIPS \cite{ying2025understanding}) and action costs $c$ (similar to \citet{zhi2024pragmatic}). SIAM efficiently inverts the rational agent model described in Section \ref{sec:synthesizing-agent-models}, computing a posterior distribution over the full sequence of latent mental states $m_{0:T}$:
{
\setlength{\abovedisplayskip}{4pt}
\setlength{\belowdisplayskip}{4pt}
\begin{multline*}
    P_\Theta(m_{0:T} | s_{0:T}, a_{1:t}) \propto P_{\theta_0}(m_0; s_0) \\ \textstyle\prod_{t=1}^T P_{\theta_m}(m_t | s_{t-1}, m_{t-1}) P_a(a_t|m_t, s_{t-1})
\end{multline*}
}
where the mental state $m_t$ can be a subset of $(g_t,b_t,r_t,c_t)$ depending on the agent model configuration $\Theta$. SIAM achieves efficiency by using incremental planning algorithms to rapidly estimate the expected utility of an action $\hat Q_{m_t}(s_{t-1}, a_t)$. We provide details in the Appendix.

Having computed a posterior over all mental states $P_\Theta(m_{0:T} | s_{0:T}, a_{1:t})$, LIRAS can answer a social inference query $Q$ by computing marginal probabilities or posterior expected values. For example, if $Q$ asks for how likely each goal is given the actions, LIRAS returns the marginal posterior $P(g_0| s_{0:T}, a_{1:t})$ over the agent's goal. If $Q$ instead asks for the cost $c_0^a$ of some action $a \in \mathsf{A}$, LIRAS returns the posterior expectation $\mathbb{E}[c_0^a|s_{0:T}, a_{1:t}]$. When $Q$ contains $k$ sub-queries, LIRAS produces $k$ corresponding ratings $R_{1:k}$ from these quantities.

\section{Experiments}
\label{sec:experiments}

\subsection{Domains}
We compare our model and baselines to human social inferences on \textbf{existing cognitive science domains} from the social reasoning literature, and a set of \textbf{expanded multimodal variants} that are derived from earlier work but that we construct specifically to evaluate the role of language in more complex, grounded environments. The existing experiments have been well-modeled by hand-constructed, domain-specific symbolic models; we choose these to assess how well our approach (and baselines) can capture judgments by synthesizing these structured models from inputs. Collectively, we choose these multi-modal domains to represent a diverse range of social reasoning tasks that vary in features including number of agents, observability, and variables of interest. 

The two existing cognitive science domains we consider are:
\begin{itemize}
    \item \textbf{Food trucks} (78 stimuli): a domain  from \cite{baker2017rational} in which the instructions describe a student navigating a campus while choosing which (movable) food truck to go to for lunch. The visual stimuli depicts varying paths taken by the agent with partial observability of the trucks. Participants are asked to jointly infer the agents’ beliefs and desires.  
    \item \textbf{Astronaut} (47 stimuli): a domain from \cite{jara2016naive} in which the instructions describe an astronaut navigating alien terrain to pick up care packages on the way to a space station. The visual stimuli show various colored terrains and paths. Participants must infer the relative costs of walking on each terrain type and the rewards of the care packages.
\end{itemize}

We also evaluate on an expanded set of multimodal domains derived from the Doors, Keys, and Gems (DKG) stimuli (introduced in \citealt{zhi2020online}, with a multi-agent version introduced in \citealt{ying2023inferring}). The original domains evaluate multi-step planning and inverse planning. Instructions describe an obstacle course in which a player is navigating a maze to reach one of several colored gems, but must first acquire various keys that unlock doors in the maze. We extend both original domains (adding significantly more stimuli to the mutli-agent case), and construct new variants that modify the linguistic instructions which specify the underlying environment dynamics:
\begin{itemize}
\itemsep0em 
    \item \textbf{DKG-Simple} (32 stimuli): 
    One colored key unlocks one door of the same color.
    \item \textbf{DKG-Double} (16 stimuli): Two colored keys unlock one door of the same color.
    \item \textbf{DKG-Reuse} (16 stimuli): One colored key unlocks all doors of the same color.
    \item \textbf{DKG-Inverse} (16 stimuli): One colored key unlocks one door of a different color.

    \item \textbf{Multiagent DKG (m-DKG)} (30 stimuli): Mazes contain two players, a principal and an assistant. The team works together to get one goal gem to the principal agent.
\end{itemize}
Each of the visual stimuli in the \textbf{DKG-Double, DKG-Reuse}, and \textbf{DKG-Inverse} variants has an identical corresponding stimulus in the base \textbf{DKG-Simple} domain, allowing paired comparison of the role of language in inferences about agent behavior.




\subsection{Human Data Collection}
For the \textbf{food truck} \cite{baker2017rational} and \textbf{astronaut} \cite{jara2016naive} domains, we evaluate on the original published data, which only contains \textit{mean} human judgments (averaged over all participants) for each inference question.  For our \textbf{extended DKG} domains, we recruit $n=20$ participants for each variant (totaling 100 participants across $5$ variants; mean age = $39.40$, $55$ female, $45$ male).  All human data collection took place over a customized web interface (see interface examples in the Appendix), where the participants first completed a tutorial and comprehension check. The experiment is approved by an IRB Board at a US University. Participants were paid \$15 USD per hour. We excluded $13$ participants who gave high likelihood scores to all options.


\subsection{LIRAS Model Configuration}
We instantiate the LIRAS model with Gemini 2.0 Flash \cite{geminiteam2025geminifamilyhighlycapable} as our base VLM for all parsing and code synthesis, and the execution of the inference by SIAM takes place on a PC. During code synthesis, we provide the VLM with a generic prompt shared across all domains and variants. This prompt includes a tutorial on SIAM syntax and primitives, and one example toy domain with accompanying parses (see Appendix). We synthesize all code with temp$=1.0$ to ensure sufficient diversity in initial synthesis, using rejection sampling until we generate a sample for each stimulus in which the full pipeline runs to completion. In our experiments we use $k=1$ samples per stimulus, as qualitatively semantic variation among models that actually compile is minimal (the fully specified stimuli in the domains we use do not ultimately suggest much uncertainty over environment dynamics -- an interesting grounds for future work).

For each stimulus, we provide the model with the full visual input and linguistic experimental setup (including instructions explaining the task, concatenated with the scene scenario and query for each stimulus) shown to human participants. We also augment the instructions to explicitly specify which actions the agent can perform, and the size of the environment grid. These same visual and linguistic specifications are used for all baselines.

\subsection{Ablations and Baselines}
We compare against the following alternatives:

\begin{itemize}
    \item \textbf{No explicit inference (ablation)}: to probe the role of the explicit Bayesian inference engine (SIAM), we run the LIRAS pipeline to fully synthesize the same symbolic environment and model representations (using the same base Gemini 2.0 Flash model), but then prompt the LLM to directly generate the answers to questions conditioned on the stimulus and generated symbolic model. 

    \item \textbf{Chain-of-thought}: we also compare our pipeline to more standard chain-of-thought \cite{wei2022chain} prompting. We evaluate on \textbf{Gemini 2.0 Flash} (our base neural model), \textbf{GPT-4o} \cite{openai2024gpt4technicalreport} (an alternate SOTA LM), and \textbf{OpenAI o3} (an explicit reasoning model). For all CoT baselines, we use temp=1 (to match our model) and report average inferences over k=3 samples to better estimate the posterior (our Appendix shows that we find significantly more variability, and worse performance, with any set of just k=1 samples.)

\end{itemize} 
We also experimented with the AutoToM \cite{zhang2025autotom} framework, which is explicitly designed for text-based Bayesian Theory of Mind. As AutoToM was only designed for natural language inputs, we experimented with first prompting a VLM (we tried both Gemini 2.0 Flash and GPT-4o) to generate a verbal narrative of the visual stimuli which could be provided to the text-based AutoToM model for end-to-end reasoning. However, we found widespread failure using either base neural model to directly generate an accurate or complete account of the visual inputs in natural language, making downstream text-based reasoning unreliable (see Appendix). 




\begin{table*}[ht!]
\small
\centering
\begin{tabular}{lccccc}
\toprule
\textbf{Models}
& \multicolumn{2}{c}{\textbf{Foodtruck}}
& \multicolumn{2}{c}{\textbf{Astronaut}}
\\
Inference & Belief & Desire
& Rewards & Cost
\\
\midrule
Gemini 2 Flash & 0.01 [-0.13, 0.15] & 0.23 [0.10, 0.35] & 0.11 [-0.15, 0.33] & 0.02 [-0.24, 0.23] \\
GPT 4o           & -0.03 [-0.18, 0.12] & 0.15 [0.02, 0.26] & -0.20 [-0.42, 0.05] & -0.17 [-0.36, 0.04]\\
OpenAI o3         & 0.52 [0.41, 0.62] & 0.45 [0.35, 0.54]  & 0.33 [0.12, 0.49]  & 0.80 [0.69, 0.88]\\ 
LIRAS (Ablated) & 0.03 [-0.10, 0.15] & 0.19 [0.06, 0.32] & 0.20 [-0.04, 0.40] & -0.08 [-0.28, 0.14]\\
LIRAS (Full)            & \textbf{0.80 [0.72, 0.86]} & \textbf{0.75 [0.67, 0.82]} & \textbf{0.87 [0.76, 0.94]} & 0.75 [0.63, 0.85] \\
\bottomrule
\end{tabular}
\caption{Correlation coefficients and corresponding 95\% confidence intervals comparing each model against human judgments on various cognitive domains for social reasoning. DKG results are displayed in Table 2. Scatterplots for each domain are shown in the Appendix.} 
\label{tab:correlation-main}
\end{table*}

\section{Results}

\begin{table*}[h!]
\small
\centering
\begin{tabular}{lccccc}
\toprule
\textbf{Models} & \textbf{DKG-Single} & \textbf{DKG-Double} & \textbf{DKG-Reuse} & \textbf{DKG-Inverse} & \textbf{m-DKG} \\
\midrule
Gemini 2.0 Flash & 0.50 [0.34, 0.64]& 0.19 [-0.10, 0.47]& 0.21 [-0.01, 0.41]& 0.12 [-0.15, 0.38]& 0.11 [-0.16, 0.37] \\
GPT 4o & 0.40 [0.24, 0.55]& 0.39 [0.07, 0.68] & 0.29 [-0.00, 0.58]& 0.11 [-0.13, 0.35]& 0.36 [0.13, 0.57]\\
OpenAI o3  & 0.73 [0.60, 0.83] & 0.73 [0.60, 0.83] & 0.52 [0.34, 0.69] & 0.79 [0.70, 0.87] & 0.81 [0.73, 0.88]\\
LIRAS (Ablated) & 0.57 [0.42, 0.69] & 0.42 [0.12, 0.66] & 0.45 [0.20, 0.66] & -0.11 [-0.31, 0.13] & 0.53 [0.33, 0.71] \\
LIRAS (Full) & 0.79 [0.70, 0.84] & 0.74 [0.58, 0.83] & 0.75 [0.61, 0.83] & 0.74 [0.61, 0.86] & 0.78 [0.70, 0.85]\\
Human (Split-half) & 0.78 [0.70, 0.84] & 0.73 [0.60, 0.84] & 0.80 [0.72, 0.88] & 0.80 [0.74, 0.86] & 0.73 [0.63,0.81] \\
\bottomrule
\end{tabular}
\caption{Correlation coefficients for each model on the four variants of the single-agent DKG domain and the multi-agent DKG domain. Overall, LIRAS shows robust correlation against humans across all variants. Other VLM baselines correlate moderately well on the DKG-Single variant, but they are less robust on some other variants with more unusual but interesting dynamics. Model results statistically significant from others are bolded. Scatterplots for each variant are shown in the Appendix.}
\label{tab:correlation-dkg}
\end{table*}

\paragraph{LIRAS demonstrates human-like reasoning on social reasoning tasks across domains.}

We compare LIRAS and baselines against human data. Table~\ref{tab:correlation-main} and ~\ref{tab:correlation-dkg} presents the correlations between model predictions and human judgments. We find that LIRAS achieves substantially higher correlation with human responses compared to many alternatives (Table~\ref{tab:correlation-main}). The Gemini 2.0 Flash vanilla model, the foundational visual and parsing component for LIRAS, shows markedly weaker performance  (and even several instances of negative correlation with human judgments). One natural question however is whether such models simply cannot reason over visual inputs. We find that the ablated LIRAS model, which synthesizes symbolic world and agent models, but instead uses an LLM to perform probabilistic inference, performs significantly worse than the full model and no better than the Gemini 2.0 Flash base model. This demonstrates the importance of the Bayesian inference engine and highlights that the failure of the smaller VLM models may be beyond visual parsing: even when given a full symbolic representation, they are unable to perform human-like probabilistic social reasoning.


While the state-of-the-art multimodal reasoning model OpenAI o3 exhibits stronger alignment with human judgments than lighter-weight models such as Gemini Flash, its performance still lags significantly behind that of LIRAS and shows a much weaker correlation against human judgments (average $r = 0.63$) on these classical cognitive social reasoning domains. These findings underscore a key limitation of contemporary vision-language models: even when extensively pre-trained and fine-tuned for complex reasoning tasks, they still face challenges in achieving human-like multimodal social reasoning, particularly when prompted to interpret visual scenes conditioned on complex linguistic information about the domain. These results show that grounded Theory-of-Mind reasoning with both language and visual inputs is challenging for most state-of-the-art foundation models -- even in classic relatively simple social reasoning domains, echoing recent findings by \citet{buschoff2025Multimodal}.

\begin{figure}[ht!]
    \includegraphics[width = 0.5 \textwidth]{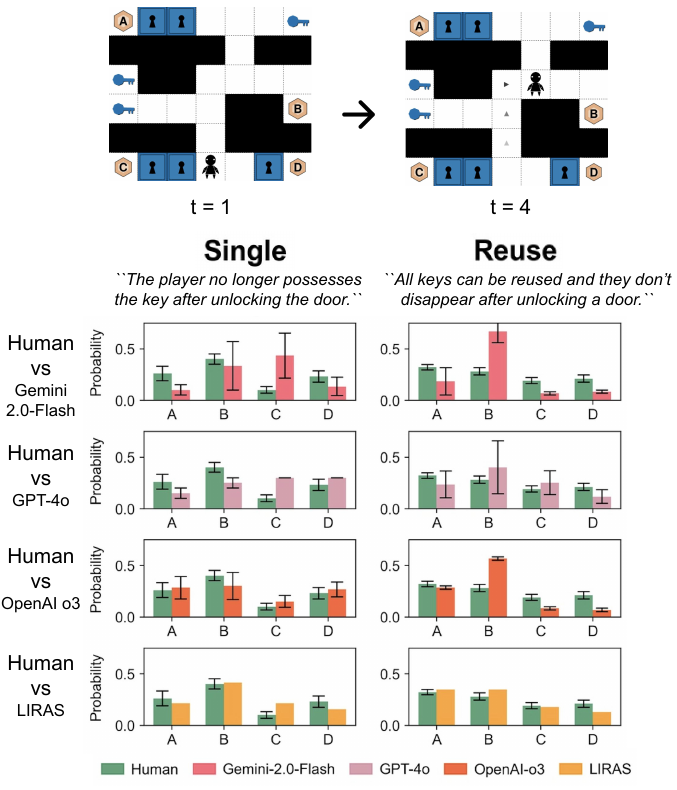}
    \caption{A qualitative example showing how models and human participants adjust their goal inference subject to changes in game dynamics. Top) Two Frames showing the agent walk a few steps up then turn right. Bottom) Model vs Human judgments on the player's goal. In the DKG-Reuse condition where each key can unlock multiple doors, humans and LIRAS find Gem A and B to be almost equally likely, where other baselines generally infer that Gem B is likely the agent's goal.}
    \label{fig:qualitative}
\end{figure}

\paragraph{LIRAS's inference is adaptive and robust to meaningful changes in linguistic inputs.}

A key motivation for this work is to computationally account for how language shapes inferences about social situations. To that end, we next assess LIRAS under variants of the DKG domain that modulate the underlying rule structure, using language. We find in Table \ref{tab:correlation-dkg} that that, similar to the model performances across domains, LIRAS performance is robust across all DKG variants, with correlations roughly at the noise ceiling of the human data (as computed under split-half correlations). We again find that ablations of LIRAS impair performance -- as do alternate VLM baselines (Gemini 2.0 Flash and GPT-4o) especially as rules become more unusual. While o3 generally achieves similar performance, we similarly notice a marked drop in performance for the DKG-Reuse condition, where a key can be reused to unlock multiple doors. 

To illustrate, we also show a qualitative example in Figure \ref{fig:qualitative}. In this example, there are four gems, A, B, C, D. The same visual stimulus were tested under DKG-Single and DKG-Reuse variants. In the Single condition, the OpenAI o3 model finds A and B to be similarly likely, where humans and LIRAS both rate gem B to be more likely. This is because the agent would have gone to get the closer blue key(s) on the left, if they are aiming for gem A, C, or D. Under the DKG-Reuse condition, where the participants are told that each key can be reused after unlocking doors, human judgments change significantly. LIRAS is able to capture this shift in probability distribution, reasoning that now the agent is \textit{equally} likely to go for gem A or gem B, since the top right blue key can now unlock \textit{both} blue doors to get to gem A. On the other hand, all other baseline models judge only gem B to be the most likely option.







\subsection{Error Analysis}

Overall after inspecting the CoT tokens by the Gemini 2.0 and GPT 4o model, we find widespread hallucination and factual errors in its reasoning. We highlight some in the appendix. The OpenAI o3 model does not output its thinking tokens. We instead ask it to justify its answer with reasoning and we also find factual mistakes and illogical statements even if the final results appear to be similar to human judgments.

In LIRAS, we also noticed some hallucination and syntax errors in world model synthesis and visual parsing. However by giving a structured prompt and parsing the grid cell by cell, LIRAS reduces the inaccuracy in the final synthesized output. Multiple checks and resampling were needed to ensure the model synthesized can be compiled and executed. 


\section{Discussion}

In this paper, we propose LIRAS, a multimodal model synthesis architecture capable of synthesizing agent and world models ``on the fly'', from visual and linguistic inputs, to reason about agents' mental states from observation. We test our model on a variety of popular cognitive science domains with different aspects of social reasoning under different settings. Like people, our model flexibly adapts to variations in the underlying rules of the scenario at hand. Our work provides a computational account of how language can help construct an ad-hoc world model for people to contextualize and interpret other agents' behavior.

On the other hand, despite using Gemini 2.0 Flash, a lightweight VLM, in the model architecture, LIRAS is able to match or outperform state-of-the-art multimodal reasoning models on most social reasoning tasks, and significantly exceeds the performance of the same Gemini 2.0 Flash model that LIRAS uses for model synthesis. By constructing these ad-hoc world and agent models on-the-fly, our study is a step towards a generalized cognitive model capable of human-like flexible social reasoning navigating the social world and form more effective human-AI thought partners~\citep{collins2024building}.



\section{Related Work}

\subsection{Model-based Theory-of-Mind Reasoning}


Our work builds on a long history of research in cognitive science and AI that shows that humans interpret others' behaviors by assuming they are rational agents \cite{dennett1981intentional, baillargeon2016psychological}. Numerous computational models have been proposed to capture this model-based reasoning process \cite{baker2017rational, shum2019theory, wu2021too, alanqary2021modeling, ying2025understanding} and have been shown to capture graded human judgments in reasoning about agents' mental states from observations.

\subsection{Theory-of-Mind in Foundation Models}

Theory-of-Mind reasoning in Foundation Models  has been subject to great interests and heated debates from the AI and NLP community. Many studies have shown that Foundational Models are capable of human-like Theory-of-Mind reasoning in many real world tasks \cite{kosinski2023theory}, while some have highlighted numerous limitations \cite{ying2025benchmarking, ullman2023large}. Recent work has also proposed cognitively inspired approach to teach LLMs to reason about linguistic social scenarios in a Bayesian way through prompting or fine-tuning~\citep{zhang2025autotom, kim2025hypothesis, qiu2025bayesian, zhu2024eliciting}. Recent work \cite{jin2024mmtom,shi2025muma} has also applied Bayesian Theory of Mind to multimodal settings by having a VLM first converting video to action predicates, although most of such existing work has focused on QA and not capturing graded human uncertainty in Theory-of-Mind reasoning.

\subsection{Automated Model Synthesis} 

The ability for Foundation Models to to synthesize code unlocks new possibilities for automated model synthesis. Automated model synthesis has been applied in different areas, from statistical reasoning~\citep{liautomated, domke2025largelanguagebayes} and planning~\cite{silver2023generalized} to cognitive modeling ~\citep{wong2023word, brooke-wilson2023bounded}. This preceding work is restricted only to language-based model synthesis, while the current work extends this to the multimodal domain. It is the first work to apply model synthesis to social reasoning, including joint synthesis over world and agent models.

\section{Limitations}

Our work is not without limitations. First, our current approach is restricted to discrete domains and does not extend to continuous spaces, reflecting a major limitation inherent in the PDDL framework. Additionally, modeling multiagent scenarios remains challenging, particularly in competitive settings; our framework cannot yet adequately capture the complexities of multiagent interactions. While we found that a single set of prompts can handle all four tested domains, the generalizability of this approach to novel domains is not guaranteed, due to possible issues with domain-specific syntactic or semantic mismatches. In addition, the LIRAS model currently parses gridworld domains by enumerating each cell, which does not generalize to more complex visual inputs. 

Lastly, our model depends on explicitly provided linguistic information—such as a clearly enumerated action space and well-defined transition dynamics (e.g., specifying that agents cannot walk through buildings). In contrast, humans can often infer such rules implicitly from context, drawing upon commonsense knowledge to build mental models of new environments without explicit instructions.

To address these limitations, future research could focus on scaling up and improving generalization by fine-tuning vision-language models (VLMs) with more diverse training examples across broader domains. This could enable more robust handling of new task domains on-the-fly. Furthermore, enhancing the model’s capacity to infer implicit domain constraints — possibly by incorporating structured priors — could make this approach more generalizable to cases where the linguistic information is ambiguous or incomplete.

\bibliography{reference}

\appendix
\onecolumn

\section{Human Data Collection}

\begin{figure*}[h]
    \centering
    \includegraphics[width=0.84\linewidth]{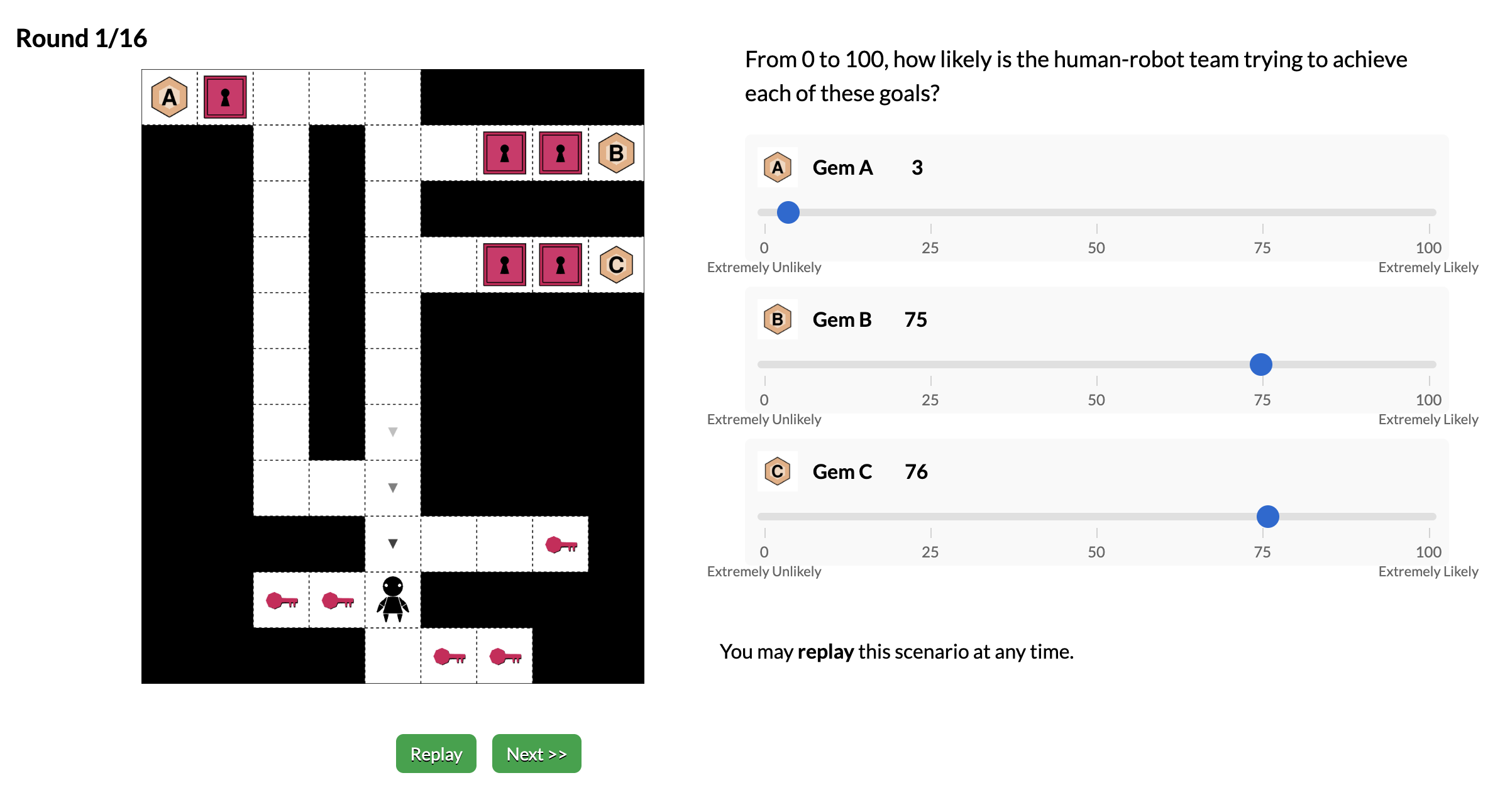}
    \caption{Experiment Interface for the human data collection.}
    \label{fig:interface}
\end{figure*}

The interface used for human data collection is shown in Fig. \ref{fig:interface}. Human participants first complete a consent form and a tutorial for the experiment. They then complete 16 trials of the study in a randomized order. In each trial, they are asked to watch the animation and then rate the likelihood for each goal gem from 0 (Extremely Unlikely) to 100 (Extremely Likely). The results are then normalized as a probability distribution across four ratings such that they sum up to 1. All human subject data collected are anonymized.

\section{Analysis of Qualitative Examples}

Snippets of Gemini 2.0 Flash reasoning on the DKG-Reuse condition for the example highlighted in Fig. \ref{fig:qualitative}.
\vspace{0.5cm}
\begin{mdframed}
**Gem-A**: Far from the starting position, requiring navigation across the locked doors. No indication of movement in that direction so far.
**Gem-B**: Accessible directly without requiring a key. The player didn't move toward it, even though it's relatively close.
**Gem-C**: Behind a door that requires a blue key. The player hasn't moved toward a key to unlock it.
**Gem-D**: Also behind a door requiring a blue key. Similar to Gem-C, no movements have been made toward unlocking it.
\end{mdframed}
\vspace{0.5cm}

Snippets of GPT 4o 2.0 reasoning on the DKG-Reuse condition for the example highlighted in Fig. \ref{fig:qualitative}.
\vspace{0.5cm}
\begin{mdframed}
 The player is likely trying to collect gem B as there is no way for the player to reach gem A or D without going to the right, and the player is heading towards that side. Alternatively, the player could be heading to get the key to unlock gem D or C.But, the key on the right is more accessible. So, it is more likely that the player is going to collect gem B.
 [5, 80, 5, 10]
\end{mdframed}
\vspace{3.5cm}
Snippets of OpenAI o3's reasoning on the DKG-Reuse condition for the example highlighted in Fig. \ref{fig:qualitative}.
\vspace{0.5cm}
\begin{mdframed}
•  Gem B (right-centre): From the current square the quickest, completely key-free way to a gem is to walk right through the top of the vertical gap (col 6) and then drop down to gem B. This route is open and costs only a few steps, so B is the most natural target.

•  Gem A (top-left): To obtain A the player must
– keep moving right to the column-6 opening,
– climb to the top row,
– walk all the way to the blue key at the far right,
– return left and unlock two blue doors, then collect A.
The present position is consistent with the first of those steps, but the plan is quite a bit longer and key-dependent.

<answer>[10,60,5,25]</answer>
\end{mdframed}

\section{Representing Spatial Input for Text-Based Reasoning Models}

We also attempted to compare LIRAS against AutoToM~\citep{zhang2025autotom} a recent work which similarly combines LLM-guided model synthesis with Bayesian inference for social reasoning. However, AutoToM is restricted to textual input. Our stimuli here critically revolve around rich visual input extended over multiple time steps. While we attempted to automatically convert the GIFs to text-only narratives describing the scene, we found that the descriptions were often impoverished and not giving AutoToM a fair chance. We show a few example narratives to demonstrate the challenges. We do so not to illustrate that current models are incapable of describing such scenarios but to highlight that baseline use for such visual description is nontrivial.

We tried both Gemini 2.0 Flash and GPT-4o. We prompted Gemini through the API, concatenating png slices of the GIF; we used the default temperature $1.0$. We separately prompted GPT-4o through the chat interface (uploading the GIF directly). In both cases, we described the Doors, Keys, Gems game environment and task the models with describing the scene in detail (where objects were and what the agent did). We depict example representative narrative descriptions for the figure we show in Figure~\ref{fig:qualitative}. 

\textbf{Gemini:} 
\begin{mdframed}
    The player begins in the middle of the screen. The player moves down one square. Then the player moves up one square.    
\end{mdframed}

\textbf{GPT-4o:}
\begin{mdframed}
The player starts near the center of the grid.
They move up one space to get a better view.
Spotting a blue key to the left, they move left and pick it up.
With the first key in hand, the player returns to the center and heads right.
They acquire a second blue key on the right side.    
\end{mdframed}

\section{Sequential Inverse Agent Modeling}

In Algorithm \ref{alg:siam}, we provide pseudo-code for the Sequential Inverse Agent Modeling (SIAM) algorithm introduced in Section \ref{sec:siam}. We show the case where all possible mental states we consider (goals, rewards, costs, and beliefs) are jointly inferred, since dropping any of the mental states corresponds to a special case.\footnote{Modulo minor differences --- when goal rewards $\mathsf{R}$ are not specified, goals are assumed to be distributed from a uniform prior. When rewards are specified, the agent is modeled as initially selecting a goal from a Boltzmann distribution over the net utility of each goal (i.e. goal reward minus shortest path cost to goal), as in \citet{jara2020naive}.} We note that since we consider deterministic PDDL environments, the expected utility of reaching a goal $g$ is equivalent to the goal reward $r_g$ plus the cost of the shortest path of reaching that goal. The cost of the shortest path \textsc{path-cost} can be efficiently computed via A* search, with the results memoized. Shortest path computations can even be incrementally computed using approaches like tree-adaptive A* \cite{hernandez2011tree}, such that once a shortest path is found from state $s$, finding a shortest path from a nearby state $s'$ is often much cheaper. We refer the reader to the appendices of \citet{ying2025understanding} and \citet{zhi2024pragmatic} for more explanation of how memoization and incremental planning can be exploited by SIPS-derived algorithms like SIAM.

\begin{algorithm}[H]
\footnotesize
\caption{Sequential Inverse Agent Modeling (SIAM) for mental state inference}
\label{alg:siam}
\begin{algorithmic}[1]
\Procedure{SIAM}{$\mathsf{G}, \mathsf{R}, \mathsf{C}, \mathsf{B}_0, a_{1:T}, s_{0:T}$}
    \State $\mathcal{H} \gets \mathsf{G} \times \mathsf{R} \times \mathsf{C} \times \mathsf{B}_0$
    \Comment{Enumerate all hypotheses (goal, reward, cost, \& belief combinations).}
    \State $\mathcal{W} \gets \{w^i := P(g^i | r^i, c^i, s_0)\}_{i=1}^{|\mathcal{H}|}$
    \Comment{Initialize (unnormalized) weights for all hypotheses.}
    \For{$t \in [1, T]$}
        \For{$h^i := (g^i, r^i, c^i, b_{0:t-1}^i) \in \mathcal{H}$}
            \State $b_t^i \gets \textsc{belief-update}(b_{t-1}^i, s_t^i, a_{t-1})$
            \Comment{Simulate agent's belief update.}
            \State $h^i \gets (g^i, r^i, c^i, b_{0:t}^i)$
            \State $Q_\text{Bel}(g^i, r^i, c^i, b_t^i, \tilde a) \gets 0$  \textbf{for} $\tilde a \in \textsc{valid-actions}(b_t^i)$
            \Comment{Initialize belief-space $Q$-values.}
            \For{$(\tilde s, \tilde w) \in b_t^i$ \textbf{and} $\tilde a \in \textsc{valid-actions}(\tilde s)$}
            \Comment{Iterate over environment states in agent's belief.}
               \State $Q^*(g^i, r^i, c^i, \tilde s, \tilde a) \gets \textsc{memoized}(\textsc{path-cost}(\tilde s, \tilde a, g^i, r^i, c^i) + r^i_{g^i})$
                \Comment{Compute shortest path cost to $g$.}
                \State $Q_{\text{Bel}}(g^i, r^i, c^i, b_t^i, \tilde a) \gets Q^*(g^i, r^i, c^i, b_t^i, \tilde a) + \tilde w \cdot Q(g^i, r^i, c^i, \tilde s, \tilde a)$
                \Comment{Update belief-space $Q$-values.}
            \EndFor
            \State $P(a_t | b_t^i, g^i, r^i, c^i) \gets \frac{\exp(\beta Q_\text{Bel}(g^i, r^i, c^i, b_t^i, a_t))}{\sum_{a} \exp(\beta Q_\text{Bel}(g^i, r^i, c^i, b_t^i, a))}$
            \Comment{Compute likelihood of action $a_t$.}
            \State $w^i \gets w^i \cdot P(a_t | b_t^i, g^i, r^i, c^i)$
            \Comment{Update weight with action likelihood.}
        \EndFor
    \EndFor
    \State \Return $(\mathcal{H}, \mathcal{W})$
    \Comment{Return all hypotheses and their (unnormalized) weights.}
\EndProcedure
\end{algorithmic}
\end{algorithm}

\section{LIRAS Model Synthesis Prompts}
\subsection{Prompt for synthesizing the PDDL domain model}

{
\scriptsize
\begin{verbatim}
"""
You will read instructions about a game setup and a dictionary of objects presented
in the problem. You will then synthesize a PDDL domain for the text description you are
given.

Note that in our PDDL definition, we use a bit-matrix and array to represent different types
of cells. These cell types generally refer to generic barriers or kinds of terrains / spaces.

If the same actions have different costs depending on the cells it is located, then each type
of cell should have a separate action definition. Note that the costs of actions on each 
terrain will be represented in a separate file and you do not need to encode that in the PDDL 
domain file. If certain cells represent barriers, then make sure you cannot move onto those 
cells.

You must include all generic_objects from the object dictionary in the PDDL types. 
The predicates should be about the states, relations or attribute of objects 
(e.g. you can define isshape, iscolor, isempty, etc.). Please do not include predicates
that are not relevant for the agents' goals (i.e. don‘t represent every possible
object attribute). The types in the PDDL domain definition should refer to broad 
category of objects (e.g. fruit) and attributes (e.g. shape, color) and not specific instances.

The task instructions will provide you with a list of actions that can be taken by the agent(s). 
Please do not invent any new actions in the PDDL domain file.

Here is an example:

Input: In this domain, you are observing a boy trying to reach some balls and plates. 
There are three unique balls: a tennis ball, a basketball and a baseball. 
The plates can have shapes of circle or square. The plates are placed inside cabinets.
The agent can move up, down, left or right. There are whitespaces and blackspace in the map.
The agents and objects can only exist in whitespaces.

objects = {'generic_objects': ['ball', 'plate', 'cabinet'], 
'unique_objects': ['tennisball', 'basketball', 'baseball'], 
'background_cells': ['whitespace', 'blackspace'],
'agent': ['boy']}

Output:

(define (domain example)
    (:requirements :fluents :adl :typing)
    (:types 
        ball plate  - item ; you may include small generic objects 
        item cabinet agent - object ; include 'item', 'agent' and any other objects
        shape ;this can be shape, color or other attributes
    )
    (:predicates 
        (has ?a - agent ?i - item)
        (at ?a - agent ?o - object) ; do not change
        (adjacent ?a - agent ?o - object) ; do not change
        (isplateshape ?p - plate ?s - shape)
        (isballshape ?b - ball ?s - shape)
    )

    (:constants 
        boy - agent ; name(s) of the agent(s) should be listed here,
        circle square - shape ; list kinds of attributes mentioned
        tennisball basketball baseball - ball ; list all unique objects
    )

    (:functions 
        (gridheight) - integer
        (gridwidth) - integer
        (xloc ?o - object) (yloc ?o - object) - integer
        (whitespace) (blackspace) - bit-matrix ; 
        this should be an exact list as in physical_generic_objects["background_cells"] 
    )

    (:derived (at ?a ?o) (and (= (xloc ?a) (xloc ?o)) (= (yloc ?a) (yloc ?o))))

    (:action pickup
     :parameters (?a - agent ?i - item)
     :precondition
        (and (not (has ?a ?i)) 
            (adjacent ?a ?i)
     :effect 
        (and (has ?a ?i)
        (assign (xloc ?i) -1) (assign (yloc ?i) -1)
        )
    )
    )

    (:action up-white
     :parameters (?a - agent)
     :precondition
        (and (> (yloc ?a) 1)
            (= (get-index whitespace (yloc ?a) (xloc ?a)) true)
            (= (get-index blackspace (- (yloc ?a) 1) (xloc ?a)) false)
        )
     :effect
        (and (decrease (yloc ?a) 1))
    )
    
    [omitting other actions]

)

In cases where you can move on the black space but at a different cost:

    (:action right-white
     :parameters (?a - agent)
     :precondition
        (and (< (xloc ?a) (gridwidth)) 
            (= (get-index whitespace (yloc ?a) (xloc ?a)) true)
        )
     :effect
        (and (increase (xloc ?a) 1))
    )

    (:action right-black
     :parameters (?a - agent)
     :precondition
        (and (< (xloc ?a) (gridwidth)) 
            (= (get-index blackspace (yloc ?a) (xloc ?a)) true)
        )
     :effect
        (and (increase (xloc ?a) 1))
    )

Multiagent cases:

    If multiple agents are present, we use an agent-code to number the agents 
    and a turn variable to indicate which agent is in turn to act.

    Include these in the functions only in cases with more than 1 agent:

            (agentcode ?a - agent) - integer
            (turn)- integer

    then you should check (= turn (agentcode ?a)) as a precondition for each agent's action. 
    Then after completing the action, we would move on to the next agent:  (assign turn (- 1 turn))


Now please generate a PDDL domain given the input below:

"""
\end{verbatim}
}

\subsection{Prompt for Parsing Image Cells}

{
\scriptsize
\begin{verbatim}
Your task is to take an image of a cell in a gridworld, then output a json file describing the object 
in the cell, using pddl, based on the pddl domain definition.

There can be multiple objects in the cell, please make sure you represent them all.

The location should be set with xloc and yloc, using placeholder $i and $j for the values. 
Please also remember to add relevant attributes such as color and shapes if applicable.

Example:

Input:

Instruction: There are pins and balls on the table. The items can be used or new. 
The pins can have a circle or square shape.


object_types: 
{
    "generic_objects":  ["pin"],
    "unique_objects": ["tennisball", "basketball", "baseball"],
    "agent": ["human"]
}

pddl_predicates: (isShape ?p - pin ?s - shape) (isNew ?i - item)


Image: [insert image of the cell showing a new circle pin and a baseball on a white square]

Output:

{
"object_name": ["pin", "baseball"]
"object_pddl_str": "(= (yloc pin) $i) \n(= (xloc pin) $j) \n(isShape pin circle) \n(isNew pin) \n(= 
(yloc baseball) $i) \n(= (xloc baseball) $j)\n "
}

if you are classifying objects that have attributes such as shapes and colors, you must include those attributes. 

\end{verbatim}
}

\subsection{Prompt for Synthesizing the Agent Configuration}

{
\scriptsize
\begin{verbatim}

Your task is to synthesize a json configuration file given the problem description. 

Grid_size: size of the grid, row by column.

Observability: should either "full" or "partial", indicating whether the agent can see the full map.

Belief config: If observability is full, return an empty dictionary, otherwise, 
include information below. Note that you should read these from 

1. Belief_object: In case of partial observability, you should indicate what is the item being hidden.

2. Belief_container: In case of partial observability, you should indicate what are the containers 
   for the hidden objects.

3. Barrier: Name any physical barrier that obstructs the view of the agent.

4. Agent: Name of the agent.

Goals: should be a list of predicate strings. If the problem doesn't call for goal inference 
(i.e. goal is given), then this should be a list with 1 predicate. Please note that sometimes the goal
can be a composite if the agent can have multiple objectives. For example, if an agent's goal is to get to get 
home, but has the option to pick up a flower or pizza on the way home, the goal space would be [["(at agent home)"], 
["(at agent home)", "(has agent flower)"], ["(at agent home)", "(has agent pizza)"], ["(at agent home)", "(has agent flower)", 
"(has agent pizza)"] ]. Make sure the predicates are allowed based on the PDDL domain definition. Make sure all goal object 
names match with the object names provided to you.

Costs: should be a list of possible different action cost profiles (dictionary). The action names should match exactly
with the actions from the PDDL description file. If the task doesn't call for different action costs (i.e. action costs vary 
across different cells), then this should have only 1 action cost profile. Action costs should be a real number greater than 0. 
In general, specific actions such as pickup should have higher costs than movements.

Query: should be one or more of the following: "belief", "goal", "reward", "cost". Note that rewards usually asks how much does agent 
like X, whereas goals questions ask which item is the agent's goal.

Temperature: How rational is the agent? The value should be a real number greater than 0. A lower temperature indicates more rational 
actions. By default this should be 1.


=========================


Here is an example:

Suppose the task is to infer a human's beliefs, goals, as well as the costs of movement in two kinds of terrains, black and 
white, at a 3 by 4 grid. The human cannot see through the black terrain.

The human's goal is to get one of the three balls: baseball, basketball, and tennisball. The balls are hidden in boxes and the agent 
cannot see which ball is in which box.

actions from PDDL domain: up-white, down-white, left-white, right-white, up-black, down-black, left-black, right-black, pickup 

object types: 
{
    "generic_objects": ["ball", "box"],
    "unique_objects": [
        "baseball", "basketball", "tennisball"
    ],
    "agent": [
        "human"
    ]
}

Example output:

{
    "grid_size": [3,4],
    "observability" : "partial",
    "belief_config" : {    
        "belief_object": "ball",
        "belief_container": "box",
        "barrier": "blackterrain",
        "agent" : "human"
    },
    "goals": [["(has human baseball)"], ["(has human basketball)"], ["(has human tennisball)"]],
    "costs": [
        {
            "up-white": 1, "down-white": 1, "left-white": 1,  "right-white": 4, "up-black": 4, "down-black": 4,
            "left-black": 4,  "right-black": 4, "pickup": 5,
    }
        {
            "up-white": 2, "down-white": 2, "left-white": 2,  "right-white": 2, "up-black": 2, "down-black": 2, "left-black":2,
            "right-black": 2, "pickup": 5,
    }
        {
            "up-white": 4, "down-white": 4, "left-white": 4,  "right-white": 4, "up-black": 1, "down-black": 1, "left-black": 1,
            "right-black": 1, "pickup": 5,
    }

    ],
    "query":["belief", "goal", "costs"],
}

Now please generate a configuration file for the following scenario:
\end{verbatim}}

\section{Scatterplots for Model and Human Judgments}

\begin{figure*}[h]
    \includegraphics[width = 1\textwidth]{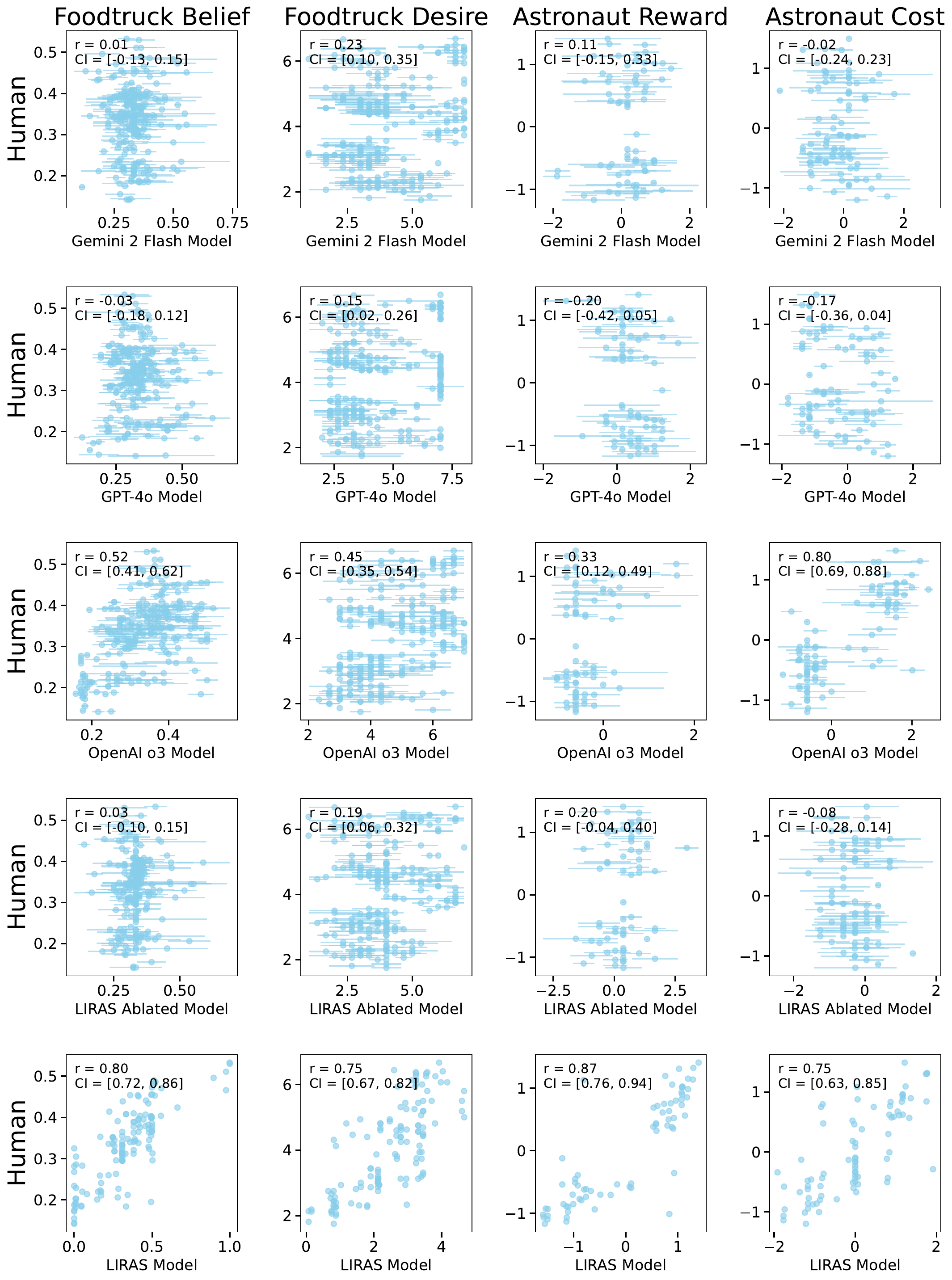}
    \caption{Scatterplots of model vs human judgments on Foodtruck and Astronaut domains. Error bar indicates standard deviation.}
    \label{fig: scatterplot-main}
\end{figure*}

\begin{figure*}[h]
    \includegraphics[width = 1\linewidth]{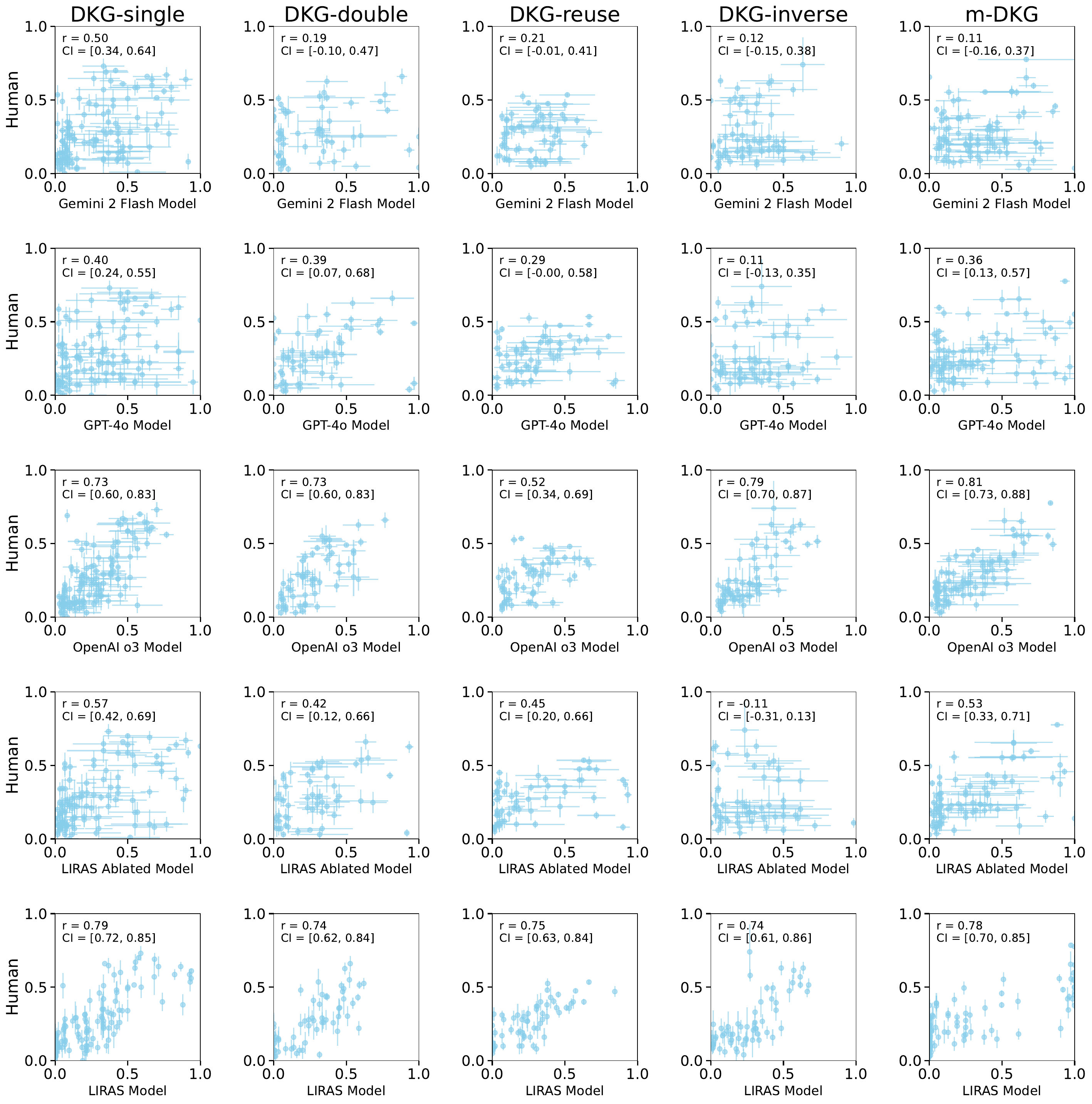}
    \caption{Scatterplots of model vs human judgments on DKG domains and variants. Error bars indicate standard deviation.}
    \label{fig: scatterplot-dkg}
\end{figure*}

\end{document}